%% file: main.tex
\documentclass{article}

\usepackage{PRIMEarxiv}

\usepackage[utf8]{inputenc} 
\usepackage[T1]{fontenc}    
\usepackage{hyperref}      
\usepackage{url}           
\usepackage{booktabs}      
\usepackage{amsmath,amsfonts,amssymb}    
\usepackage{nicefrac}     
\usepackage{microtype}      
\usepackage{lipsum}
\usepackage{fancyhdr}       
\usepackage{graphicx}     
\graphicspath{{media/}}     
\usepackage{caption}
\usepackage{orcidlink}

\usepackage{algorithmic}
\usepackage{array}
\usepackage[caption=false,font=normalsize,labelfont=sf,textfont=sf]{subfig}
\usepackage{textcomp}
\usepackage{stfloats}
\usepackage{verbatim}
\usepackage{cite}
\usepackage{bm}
\usepackage{multirow}
\usepackage{adjustbox}
\usepackage{hyperref}
\usepackage[ruled]{algorithm2e}
\usepackage{xcolor}
\hypersetup{
  colorlinks   = true, 
  urlcolor     = blue, 
  linkcolor    = blue, 
  citecolor   = red
}

\title{\emph{Diffexplainer}: Towards Cross-modal Global Explanations with Diffusion Models}

\author{
        \href{https://orcid.org/0000-0002-6721-4383}{\includegraphics[scale=0.06]{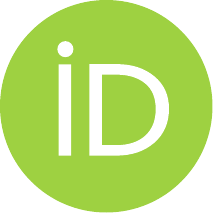}\hspace{1mm}Matteo Pennisi}\\
        PeRCeiVe Lab \\
        University of Catania, Italy\\ 
        \And
        \href{https://orcid.org/0000-0002-1333-8348}{\includegraphics[scale=0.06]{orcid.pdf}\hspace{1mm} Giovanni Bellitto}\\
        PeRCeiVe Lab \\
        University of Catania, Italy\\ 
        \And
        \href{https://orcid.org/0000-0002-2441-0982}{\includegraphics[scale=0.06]{orcid.pdf}\hspace{1mm}Simone Palazzo}\\
        PeRCeiVe Lab \\
        University of Catania, Italy\\ 
        \And
        \href{https://orcid.org/0000-0001-6172-5572}{\includegraphics[scale=0.06]{orcid.pdf}\hspace{1mm}Mubarak Shah}\\
        CRCV\\
        University of Central Florida, United States\\ 
        \And
        \href{https://orcid.org/0000-0001-6653-2577}{\includegraphics[scale=0.06]{orcid.pdf}\hspace{1mm}Concetto Spampinato}\\
	PeRCeiVe Lab \\
        University of Catania, Italy
}

\input{macros}

\begin{document}
\maketitle

\begin{figure}[htb!]
  \centering
  \vspace{-20pt}
  \includegraphics[width=1\textwidth]{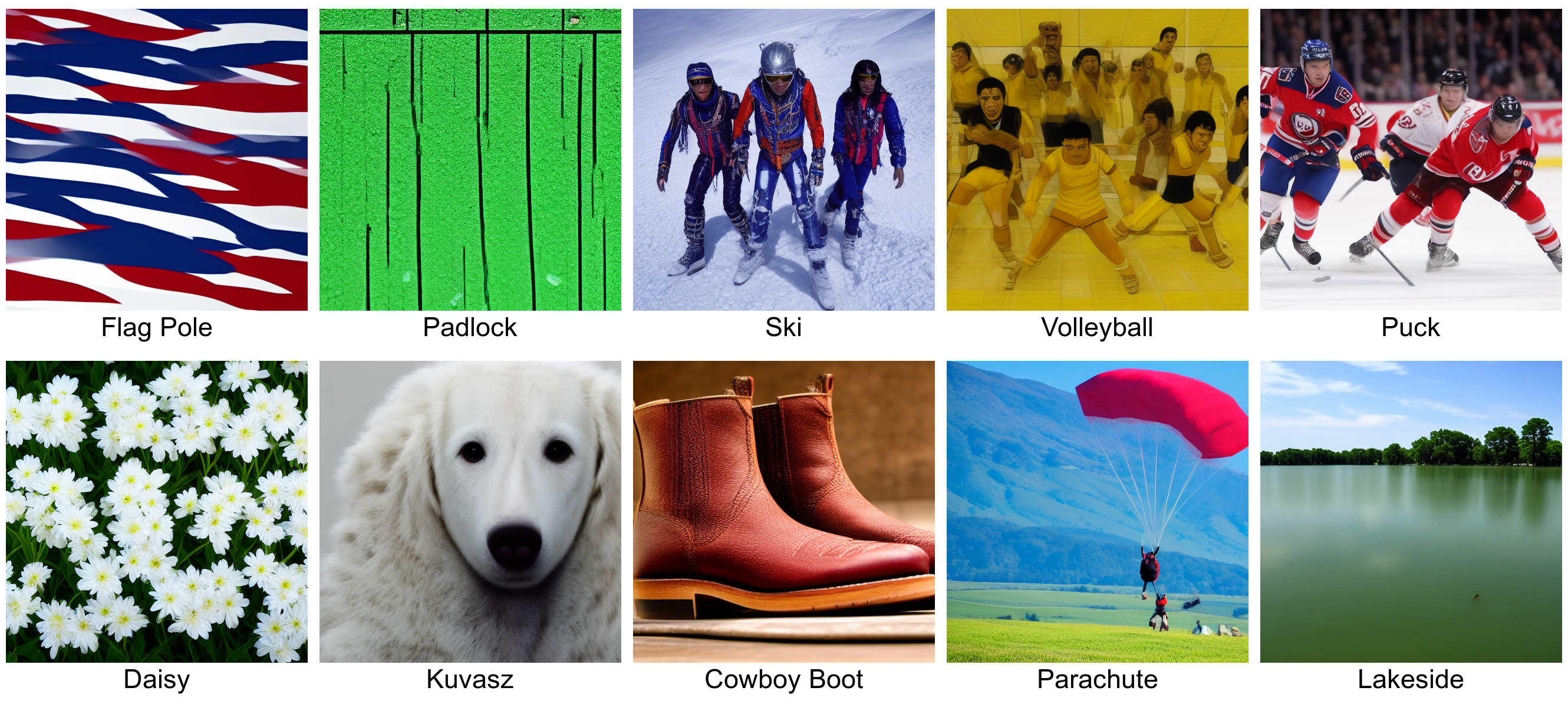}
  \caption*{\textbf{DiffExplainer generates images that maximize the activation of a target classifier} (Robust ResNet-50 trained on ImageNet in the examples). Top: biased cases, where object categories are missing from the images that maximize class outputs (e.g., ``Padlock'' contains a pattern of wood but no padlock is present). Bottom: unbiased cases, where the target class is depicted in the generated images.}
  \label{fig:abstract}
\end{figure}

\begin{abstract}
\input{sections/abstract}
\end{abstract}


\input{sections/introduction}
\input{sections/related}
\input{sections/method}
\input{sections/results}
\input{sections/conclusions}
\newpage


\end{document}

%% file: macros.tex
\newcommand{\mytilde}{\raise.17ex\hbox{$\scriptstyle\mathtt{\sim}$}}
\DeclareMathOperator*{\argmin}{arg\,min}
\DeclareMathOperator*{\argmax}{arg\,max}

\def\cC{\mathcal{C}}
\def\dD{\mathcal{D}}
\def\eE{\mathcal{E}}

\def\tT{\mathcal{T}}

\def\xX{\mathcal{X}}
\def\yY{\mathcal{Y}}
\def\zZ{\mathcal{Z}}

\def\Ee{\mathbb{E}}

\def\Re{\mathbb{R}}

\def\Te{\mathbb{T}}

%% file: sections/abstract.tex
We present \textit{DiffExplainer}, a novel framework that, leveraging language-vision models, enables multimodal global explainability. DiffExplainer employs diffusion models conditioned on optimized text prompts, synthesizing images that maximize class outputs and hidden features of a classifier, thus providing a visual tool for explaining decisions. Moreover, the analysis of generated visual descriptions allows for automatic identification of biases and spurious features, as opposed to traditional methods that often rely on manual intervention. The cross-modal transferability of language-vision models also enables the possibility to describe decisions in a more human-interpretable way, i.e., through text. We conduct comprehensive experiments, which include an extensive user study,  demonstrating the effectiveness of DiffExplainer on 1) the generation of high-quality images explaining model decisions, surpassing existing activation maximization methods, and 2) the automated identification of biases and spurious features.

%% file: sections/introduction.tex
\section{Introduction}
\label{sec:introduction}
In recent years, the proliferation of deep learning systems across various domains has revolutionized many aspects of society, from healthcare to transportation. These systems have demonstrated remarkable capabilities in tasks ranging from image classification to natural language processing. However, alongside their widespread adoption comes a pressing need to ensure the trustworthiness, explainability,  reliability, and fairness of these models, especially in high-stakes applications where erroneous decisions can have significant consequences.

One of the critical challenges undermining the reliability of deep learning models is their susceptibility to spurious input features, i.e., features that are unrelated to the underlying task but influence model predictions. 
This reliance on superficial cues can lead to unexpected and erroneous behavior, affecting the integrity and trustworthiness of the models. For instance, a deep neural network trained to identify objects in images may inadvertently base its decisions on irrelevant features such as background color or texture rather than the object's inherent characteristics\cite{singla2022salient, geirhos2018imagenettrained,sagawa2019distributionally,koh2021wilds,santurkar2021breeds,pmlr-v206-ye23a}.

Addressing these challenges requires new methods able to identify, explain and  mitigate biases in order to improve transparency and reliability. 
Traditional methods for identifying spurious features often rely on manual intervention, which is labor-intensive, time-consuming, and impractical for large-scale datasets. The recent Salient ImageNet dataset~\cite{singla2022salient} describes 
one of these attempts, revealing that popular ImageNet-based models, such as ResNet-50, heavily rely on various spurious features in their predictions.
Furthermore, existing techniques (e.g., activation maps~\cite{8237336}, feature attacks~\cite{Singla_2021_CVPR}, etc.) for explaining model decisions often fall short in providing comprehensive insights or unambiguous representations  of the underlying reasoning processes, limiting their utility in critical applications where transparency is paramount.

We propose a novel framework, DiffExplainer, that employs a language-conditioned diffusion model to synthesize images that maximize the responses of individual neurons of a visual classifier, by performing gradient descent on the text embedding space. Leveraging the cross-modal transferability of language-vision models grants us several advantages. First, we harness the generation capabilities of state-of-the-art text-to-image models to produce clear and recognizable explanations of high-level concepts. Second, we are able to employ textual cues, such as class labels, to verify that the activations within the model correspond to the anticipated semantic visual concepts; in turn, this supports the identification of spurious features and biases learned by the model. Third, our approach enables targeted investigations by crafting the textual prompts, thereby exerting control over the generation of specific textures and shapes. Finally, operating in the joint text-image space opens the possibility to provide human-interpretable textual explanations for model decisions. 

We demonstrate the validity of our approach through comprehensive experiments, showcasing its ability to generate high-quality images that explain model decisions and to identify spurious features without human supervision. Additionally, we present the results of a user study providing qualitative and quantitative evidence of the effectiveness of our framework. 
Finally, DiffExplainer reveals failures in Salient ImageNet, due to local analysis (i.e., activation maps shown to users), while providing at the same time a more thorough explanation (including information about shape, context, texture) of model decisions on ImageNet.

%% file: sections/related.tex
\section{Related Work}
\label{sec:related}

Our work, which is rooted in human-interpretable explanation and on identifying spurious correlations, is situated within a broader context of research aiming to enhance model transparency and interpretability. In this section, we delineate three categories of related work: activation maximization, interpretation and discovering spurious correlations.

\noindent \textbf{Model interpretation}. The landscape of model interpretation encompasses a variety of methods, including attribution-based approaches~\cite{10.1145/2939672.2939778,NIPS2017_8a20a862,10.5555/3305890.3306006} and concept-based methodologies~\cite{NEURIPS2019_77d2afcb,pmlr-v119-koh20a}. While these techniques contribute to elucidating the model prediction process, their outputs often present complexities for human comprehension and show inconsistencies across models and algorithms~\cite{NEURIPS2019_77d2afcb,jain2022missingness}. Moreover, some methodologies require modifications to model architectures or entail complex post-processing steps. The most used interpretation methods, e.g., GradCam~\cite{8237336} or Integrated Gradients~\cite{integrated_gradients}, provide detailed insights into individual predictions, but their limited scope and sensitivity to perturbations may hinder generalizability. Conversely, global post-hoc explanations offer a broader understanding of model behavior but may sacrifice instance-specific peculiarities. Our work adopts a novel approach to interpreting the model prediction process by leveraging language through pre-trained diffusion models to identify the most influential attributes. This framework facilitates the derivation of meaningful interpretations without the need for extensive pre-processing or post-processing steps. 

\noindent \textbf{Activation maximization}. A common strategy to provide global explanations of model decisions is by delving into neuron behavior to identify their preferred stimuli~\cite{YosinskiCNFL15,zeiler2014}. Originally, preferred inputs were sought from extensive image collections, such as training or test sets~\cite{zeiler2014}. However, evaluating each neuron across a vast array of images requires significant computational demands, compounded by the challenge of missing informative images and the non-trivial interpretation of specific visual features within images. Activation maximization, i.e., the automatic synthesis of preferred stimuli, has garnered attention~\cite{NguyenYC15,NguyenDYBC16,NguyenYC16,SimonyanVZ13,YosinskiCNFL15,Mahendran16} to address the above limitation. This process typically starts with a random image and iteratively refines it through backpropagation to enhance the activation of a target neuron. However, previous studies have revealed that unconstrained optimization can yield unrealistic and uninterpretable images, as the vast space of possible images includes ``fooling'' inputs that excite a neuron without resembling natural images~\cite{NguyenYC15}. To address this limitation, optimization methods constrain synthetic images to closely resemble natural images, by incorporating specific priors into the objective function~\cite{SimonyanVZ13,Mahendran16,YosinskiCNFL15}. One strategy to learn realistic priors has been proposed by Nguyen et al.~\cite{NguyenDYBC16}, where a deep generator is employed as an upstream module to learn visual priors of realistic images. This approach is the most similar to our work, but presents a major drawback, in that it is specifically confined to generating samples solely from the GAN's trained data distribution, constraining the search for biases to those encoded in the latent space and hindering the discovery of new biases. DiffExplainer, instead, offers enhanced control over image synthesis by operating in a text embedding space, enabling finer adjustments compared to methods limited to visual input, thus supporting the exploration of new visual patterns.
Examples substantiating this claim are reported in Section~\ref{sec:Results}.

\noindent \textbf{Discovering spurious features}. Numerous studies have investigated spurious correlations --- features that are not essential to object definition but often co-occur with the object --- across various domains, including image textures, backgrounds, domain shifts, and causally unstable attributes~\cite{geirhos2018imagenettrained,sagawa2019distributionally,koh2021wilds,santurkar2021breeds,pmlr-v206-ye23a}. Approaches to detecting spurious correlations range from leveraging domain knowledge to inferring spurious attributes through environment learning~\cite{ClarkYZ19,creager2021environment,10.5555/3495724.3497346,Seo_2022_CVPR}. Recent works also employ explainability techniques for identifying spurious attributes, often requiring human inspection~\cite{plumb2022finding,Abid2021Meaningfullydebuggingmodel}. 
Singla et al. recently introduced the Salient ImageNet dataset~\cite{singla2022salient}, presenting a framework aimed at uncovering and localizing spurious and core visual features utilized in model inference. The Salient ImageNet approach employs activation maps of neural features as soft masks to highlight spurious or core visual characteristics, with the final discrimination between these features being conducted under human supervision.

In contrast, DiffExplainer adopts realistic examples to explain model features instead of relying on activation maps or feature attacks, as observed in prior works~\cite{Singla_2021_CVPR}. Our empirical findings demonstrate that this alternative strategy is more effective in discerning core and spurious features, as substantiated by an extensive user study. Furthermore, leveraging visual analysis of generated samples, through established foundation models, allows us to automatically infer the presence of bias, thereby enabling the application of our approach to arbitrary datasets without requiring human supervision.

%% file: sections/method.tex
\section{Method}
\label{sec:method}

\begin{figure}[tb]
  \centering
  \includegraphics[width=0.98\textwidth]{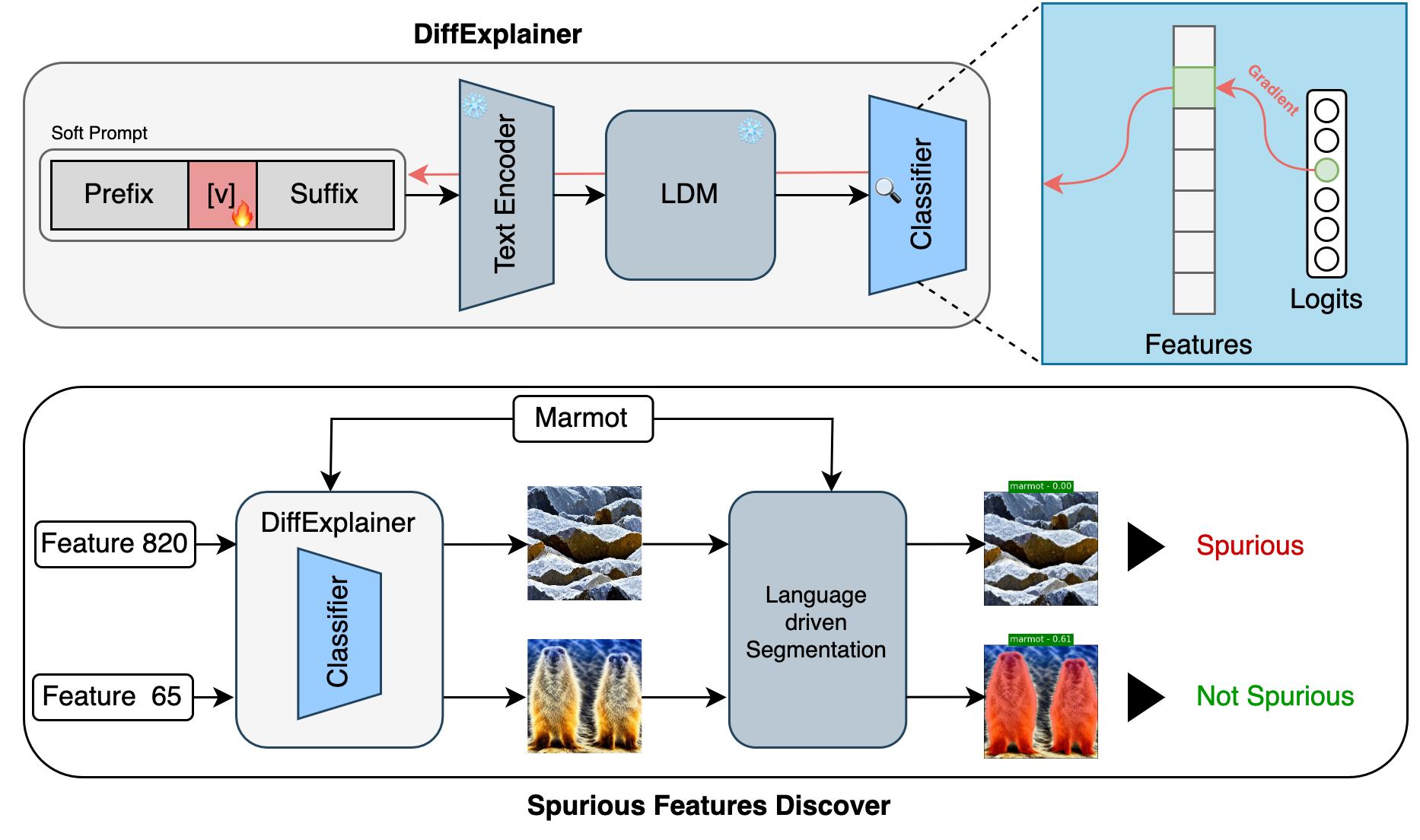}
  \caption{\textbf{DiffExplainer architecture.} Top: an optimized soft prompt is fed to a text encoder for conditioning a LDM; in turn, this generates images that maximize a specific hidden feature or the output of a classifier. Bottom: automated discovery of spurious features based on the pixel-level segmentation of synthesized images.}
  \label{fig:method}
\end{figure}

Let $\phi: \xX \rightarrow \cC$ be a visual classification model, trained to map images from $\xX$ to one of the classes in $\cC$. Let also $\phi_j: \xX \rightarrow \Re$ denote the response a specific neuron $j$ in the model, for a given visual input. Our objective is to investigate, in a broad sense, what the model has learned and the criteria it adopts to provide class decisions. 
Common approaches for model explainability translate this goal into an optimization task in the image space, seeking input samples $\hat{x} \in \xX$ that maximally activate a specific intermediate neuron $\phi_j$ or that minimize the cross-entropy loss $\text{\emph{CE}}$ for a target class $c \in \cC$, i.e.:
\begin{equation}
\hat{x} = \argmax_x \bigl[ \phi_j \left( x \right) - R(x) \bigr]~~~\text{or}~~~\hat{x} = \argmin_x \bigl[ \text{CE} \bigl( \phi \left( x \right), c \bigr) + R(x) \bigr],
\end{equation}
with $R(x)$ being an optional regularizer or realism prior to be minimized.

While our framework also supports this interpretation of model explainability, we intend to dig deeper into the semantics of what the model has learned, by operating in a shared space bridging the textual and visual domains, thus facilitating both generation and interpretation. In this setting, assuming the availability of a text-to-image model $G: \tT \rightarrow \xX$, we pose model explainability as an optimization task in the text embedding space $\tT$, with the objective of seeking textual representations $\hat{t} \in \tT$ that produce visual samples activating specific model features or output classes, i.e.:
\begin{equation}
\hat{t} = \argmax_t \phi_j \bigl( G \left ( t \right) \bigr) ~~~ \text{or} ~~~
\hat{t} = \argmin_t \text{CE} \Bigl( \phi \bigl( G \left( t \right) \bigr), c \Bigr).
\end{equation}

Note that this formulation does not require the explicit definition of a regularizer $R(x)$, as it relies on $G$'s intrinsic encoding of such realism prior. An overview of the proposed approach is illustrated in Fig.~\ref{fig:method} (top); in the following, the individual components that make up our framework are described in detail.

\subsection{Latent Diffusion Models}

Our method is implemented on top of Latent Diffusion Models (LDMs)~\cite{Rombach_2022_CVPR}, a particular type of Denoising Diffusion Probabilistic Models (DDPMs)~\cite{NEURIPS2020_4c5bcfec} where the denoising process takes place in the latent space of an auto-encoder. Formally, LDMs are made up of two components: the auto-encoder and a diffusion model. The encoder $\eE: \xX \rightarrow \zZ$ of the auto-encoder learns to map images to a latent space $\zZ$, while the decoder $\dD: \zZ \rightarrow \xX$ reverses the transformation, so that $x \approx \dD \bigl( \eE \left( x \right) \bigr)$, $x \in \xX$. 

The diffusion model $\epsilon: \zZ \times \yY \rightarrow \zZ$ is trained to produce valid codes by iteratively denoising a latent representation, starting from random values from a prior distribution\footnote{For simplicity of notation, we indicate the entire diffusion process as $\epsilon$, though in the original paper it refers to a single denoising iteration.} and controlling the entire process through a condition $y \in \yY$, which is injected into the denoiser network by means of a cross-attention mechanism. In our setting, we specifically condition the diffusion model via textual representations provided by a CLIP~\cite{radford2021learning} text encoder $\tau: \Te^N \rightarrow \tT$, thus $\yY = \tT$, with $\Te^N$ being the space of $N$-long sequences of token embeddings (in practice, $N$ is arbitrary, thanks to the properties of transformer-based text encoders). 

\subsection{Explanation Synthesis}

We formulate our explainability approach as the optimization of a sequence of textual tokens which conditions the generation of a visual sample, in turn activating specific portions of a classification model (e.g., an intermediate feature or a class output). Formally, our objective is to optimize a sequence $t = \left[ t_1, t_2, \dots, t_N \right]$, where each $t_i \in \Te$.

The construction of $t$ is arbitrary: in our base formulation, we follow a ``soft prompt'' approach, where some of the tokens in $t$ are fixed and correspond to elements of the text encoder's vocabulary, while others are variable and subject to optimization. This ensures that the textual input complies with the requirements of the text encoder $\tau$, e.g., the presence of \texttt{SOS} (start of sequence) and \texttt{EOS} (end of sequence) tokens. Given this constraint, a minimalist version of $t$ consists in $t = \left[\text{\texttt{SOS}}, t_1, \text{\texttt{EOS}} \right]$, with $t_1$ being a single optimizable token. This formulation also allows us to control the image generation process in an intuitive way: for instance, setting the fixed prompt to the string ``the shape of'' enables the investigation of what kind of shapes mostly activate some features of classifier $\phi$.

Once the structure of $t$ is defined, our goal is to find the optimal values for learnable tokens. Formally, in order to generate an explanation of a class output $c$ in $\phi$, the optimization problem can be defined as follows:
\begin{equation}
t^* = \argmin_t \Ee_{z \sim p(\zZ)} \Biggl[ \text{CE} \Biggl( \phi \biggl( \dD \Bigl( \epsilon \bigl( z, \tau \left( t \right) \bigr) \Bigr) \biggr) , c \Biggr) \Biggr],
\label{eq:argmin_class}
\end{equation}
where $p(\zZ)$ is the prior distribution on the LDM's latent space.

When optimizing for the activation of a hidden neuron $\phi_j$, we adapt the objective function as:
\begin{equation}
t^* = \argmax_t \Ee_{z \sim p(\zZ)} \Biggl[  \phi_j \biggl( \dD \Bigl( \epsilon \bigl( z, \tau \left( t \right) \bigr) \Bigr) \biggr)  \Biggr].
\label{eq:argmin_feat}
\end{equation}

Given that all components of the proposed approach are differentiable, the most natural way to solve Eq.~\ref{eq:argmin_class} and~\ref{eq:argmin_feat} is through gradient descent. The main challenge in doing so lies in the iterative nature of the latent diffusion process, which may require hundreds of step to converge, causing a significant computational burden. To overcome this difficulty, we adopt Latent Consistency Models (LCMs)~\cite{luo2023latent,luo2023lcm}, that speed up the diffusion process by treating it as an augmented probability flow ODE problem, resulting in efficient synthesis with few (less than 10) inference steps.

Another critical aspect of the optimization procedure is enforcing independence of $t^*$ from the initial diffusion noise $z$. Indeed, optimizing for a fixed $z$ leads the soft prompts in $t^*$ to overfit, thus losing the capability to synthesize multiple samples by varying the inputs noise. For this reason, Eq.~\ref{eq:argmin_class} and~\ref{eq:argmin_feat} seek a value of $t^*$ that optimizes the expectation over $z \sim p(\zZ)$. In practice, we achieve this property by randomly sampling a new $z$ at each optimization step, thus ensuring the generalizability of the optimized $t^*$.

The result of the optimization procedure is twofold. The immediate product is the optimized prompt $t^*$. It should be noted that the ``soft prompt'' approach, while being easier to optimize, does not impose a strict correspondence of optimized tokens to actual language tokens, which would require to handle the discrete nature of the vocabulary. A variant of our approach where such correspondence is indeed enforced is discussed in Section~\ref{sec:text_expl}. A secondary output is an \emph{explanation distribution} $x_{t^*}(z) = \dD \Bigl( \epsilon \bigl( z, \tau \left( t^* \right) \bigr) \Bigr)$, which can be sampled to qualitatively understand the visual concepts encoded by specific model features. In Section~\ref{sec:Results}, we present experimental results that demonstrate the suitability of the synthesized images in inspecting the model's behavior.

\subsection{Spurious Feature Discovery}
\label{sec:spurious_discovery}

Our approach for the discovery of core/spurious features combines the above explanation synthesis method with language-driven segmentation techniques to uncover possible biases.

For each class $c \in \cC$ of classifier $\phi$, we first identify the top-5 features $\left\{ f_{c,1}, \dots, f_{c,5} \right\}$ of the last fully-connected layer before the classification layer, following the procedure in~\cite{singla2022salient}, and synthesize the corresponding visual explanations. In detail, for a given feature $f_{c,i}$, we first compute $t_{c,i}^*$ by optimizing the following adapted objective (discussed below):
\begin{equation}
t_{c,i}^* = \argmin_t \Ee_{z \sim p(\zZ)} \Biggl[ \text{CE} \Biggl( \phi \biggl( \dD \Bigl( \epsilon \bigl( z, \tau \left( t \right) \bigr) \Bigr) \biggr) , c \Biggr) - \phi_{f_{c,i}} \biggl( \dD \Bigl( \epsilon \bigl( z, \tau \left( t \right) \bigr) \Bigr) \biggr)  \Biggr] ,
\label{eq:argmin_spur}
\end{equation}
and then synthesize the corresponding image by sampling from the distribution $x_{c,i}(z) = \dD \Bigl( \epsilon \bigl( z, \tau \left( t_{c,i}^* \right) \bigr) \Bigr)$. We then feed the generated images to a language-driven semantic segmentation model, conditioned by the class name (as text) to identify pixels in the image belonging to a target concept. The language-driven segmentation pipeline employs DINO~\cite{liu2023grounding} for bounding box estimation from a text prompt (the class name, in our case), followed by SAM~\cite{Kirillov_2023_ICCV}, that employs the bounding boxes as prompts for predicting segmentation masks. Given the output mask segmentation mask for the target class, we can compute a value $r_{c,i}$ as the fraction of segmented object pixels out of the entire image. The decision on whether feature $f_{c,i}$ is a core feature or a spurious feature for class $c$ is made based on the expected value of $r_{c,i}$ (which is a random variable depending on $z$, as it is computed on the segmentation output of $x_{c,i}(z)$):
\begin{equation}
\text{feature}~f_{c,i}~\text{is}~
\left\{
\begin{aligned}
&\text{core} & \text{if}~\Ee_{z \sim \zZ} \left[ r_{c,i} \right] \ge \delta \\
&\text{spurious} & \text{if}~\Ee_{z \sim \zZ} \left[ r_{c,i} \right] < \delta
\end{aligned}
\right.,
\end{equation}
with $\delta$ being a threshold hyperparameter. In practice, we approximate the expectation over $z$ by sampling 10 images from $x_{c,i}(z)$.

It is critical to note that the optimization objective in Eq.~\ref{eq:argmin_spur} differs from the one for hidden feature optimization previously presented in Eq.~\ref{eq:argmin_feat}. Specifically, the formulation in Eq.~\ref{eq:argmin_spur} optimizes the combination of Eq.~\ref{eq:argmin_class} and Eq.~\ref{eq:argmin_feat}. The reason for including the constraint on class activation in Eq.~\ref{eq:argmin_spur} is related to the consideration that a given feature in a deep fully-connected layer is likely to contribute to the activation of multiple classes; hence, a visual explanation of that feature by our method does not guarantee that an instance of the target class appears in the synthesized sample, which is required by our segmentation-driven approach for spurious feature identification. On the other hand, it is important to understand that including the constraint from Eq.~\ref{eq:argmin_class} does not guarantee either that an actual instance of the target class appears in the synthesized image: indeed, if the feature is spurious, we expect the target class \emph{not} to be present in the synthesized image, since the presence of a spurious feature itself, by definition, is enough for the model to predict the corresponding class.

%% file: sections/results.tex
\section{Experimental Results}
\label{sec:Results}
In the following section we show the visualization capabilities of DiffExplainer when compared to activation maximization methods (Section~\ref{sec:qualitative}) and its ability to discover spurious features without human intervention (Section~\ref{sec:spurious}). Furthermore, we report the results on a user study (Section~\ref{sec:user_study}) conducted in order to quantify these capabilities. 

As our primary dataset, we adopt a subset of Salient ImageNet, containing human annotations of the top-5 features (labeled as spurious features or core features) of a Robust ResNet-50~\cite{singla2022salient} trained on ImageNet. The subset is chosen to be balanced w.r.t. the number of spurious features identified among the top-5 features of each class: in detail, we select 15 classes for each level of spuriousness (ranging from 0, where none of the top-5 features are spurious, to 5, where all of them are), leading to a total of 90 classes.\\
We also report some results on discovering injected ethnicity bias  on the FairFaces~\cite{Karkkainen_2021_WACV} dataset in Section~\ref{sec:gender}.
Training procedure details and an extended version of the results can be found in the supplementary materials.
\subsection{Qualitative comparison with Activation Maximization Methods}
\label{sec:qualitative}
\begin{figure}[htb!]
  \centering
  \includegraphics[width=0.8\linewidth]{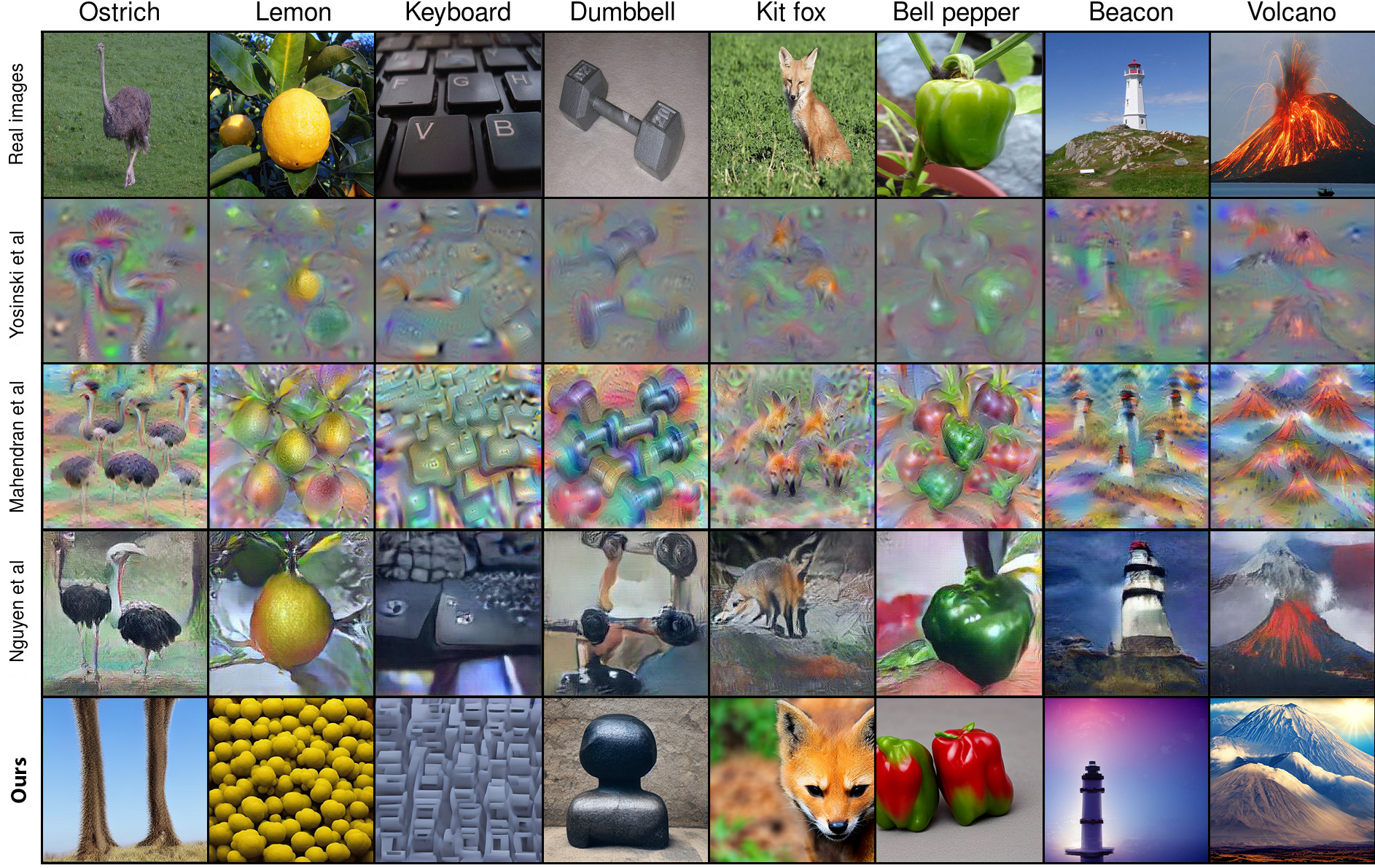}
  \caption{Visual comparison of activation maximization strategies: Yosinski et al.~\cite{yosinski2015understanding},
  Mahendran et al.~\cite{Mahendran16},
  Nguyen et al.~\cite{NguyenDYBC16} for a CaffeNet~\cite{caffenet} trained on ImageNet. DiffExplainer excels in generating higher-quality images compared to competitors.}
  \label{fig:visual_comparison}
\end{figure}

We first conduct a visual comparison between the images generated by our approach and those produced by traditional activation maximization methods. We analyze the results on a CaffeNet architecture~\cite{caffenet} to maintain consistency with previous studies. 
For this evaluation, we employ DiffExplainer to maximize the output class of the target classifier (i.e., optimizing Eq.~\ref{eq:argmin_class}). Fig.~\ref{fig:visual_comparison} illustrates the visual comparison between our generated images and those obtained through activation maximization methods.

\begin{figure}[htb!]
  \centering
  \includegraphics[width=0.98\linewidth]{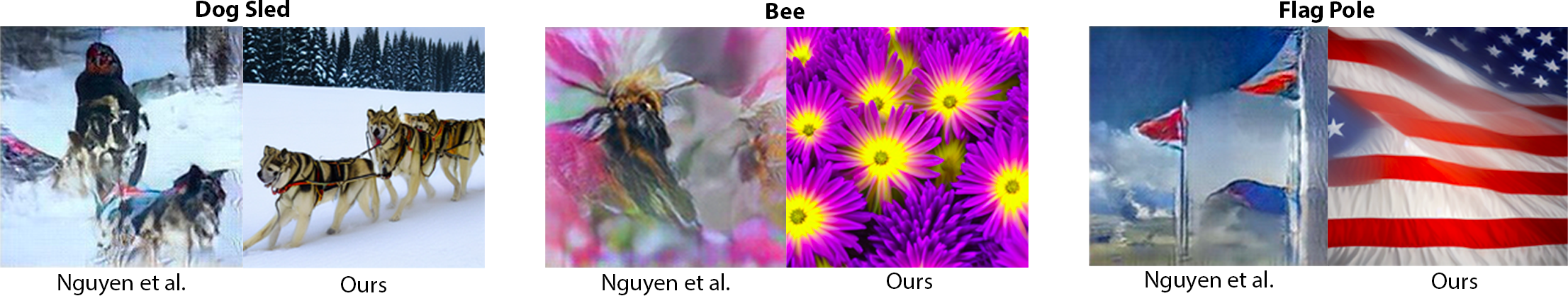}
  \caption{\textbf{Discovering classifier bias}: 
  generated samples for the CaffeNet classifier, for ``dog sled'', ``bee'' and ``flagpole'' classes. Nguyen et al.'s method~\cite{NguyenDYBC16} produces images from the latent GAN distribution, which inherently reflect the dataset used for training rather than the features used to classify. Our approach, instead, generates samples where the main category is absent, thereby elucidating specific biases: ``dog sled'' is discerned through the presence of dogs, snow, and trees; ``bee'' through the presence of flowers; and ``flagpole'' through the presence of the US flag.}
  \label{fig:nguyen_bias_final}
\end{figure}

DiffExplainer excels in generating higher-quality images compared to competitors. However, it is crucial to note that our method aims not only to enhance the quality of activation maximization images, as evidenced by our results, but also to better understand what the model has learned. For instance, for the ``Ostrich'' class, DiffExplainer synthesizes an image emphasizing ostrich legs, indicating a distinct classifier focus. In contrast, Nguyen et al.'s method~\cite{NguyenDYBC16} reconstructs an image featuring ostriches. While this might seem advantageous for \cite{NguyenDYBC16} over DiffExplainer, it actually underscores its inherent shortcoming. Indeed, the generation capabilities of~\cite{NguyenDYBC16} are limited to the visual information cast in the input latent space.
The implication is that Nguyen's method operates within the constraint of generating samples solely from the data distribution utilized during GAN training. This constraint, in turn, limits the scope of exploration concerning biases towards visual patterns that may activate output neurons, but are not explicitly encoded within the GAN latent space.

In contrast, DiffExplainer, by leveraging text embeddings, enables the model to traverse a broader semantic space that is not constrained by the visual patterns learned during GAN training. This flexibility allows DiffExplainer to uncover and illustrate biases and associations within the model that might not be directly related to the target class but are nevertheless influential in the model's decision-making process. 
For instance, when optimizing the text embedding to maximize the ``bee'' class, the embedding shifts closer to ``flower'', resulting in the generation of flower images and hinting on the presence of a bias in the dataset. Conversely, \cite{NguyenDYBC16} persists in generating bee images, potentially concealing underlying biases. Examples of the different behaviors between the two methods are shown in Fig.~\ref{fig:nguyen_bias_final}.
\begin{figure}[hbt!]
  \centering
  \includegraphics[width=0.98\linewidth]{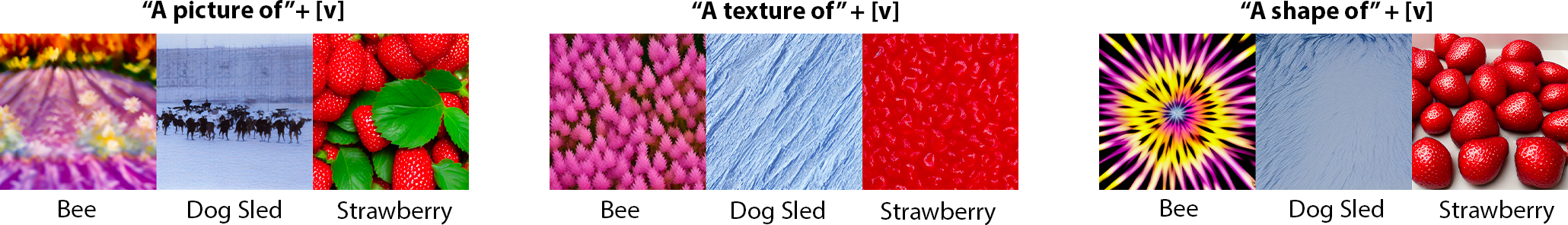}
  \caption{Examples showcasing the benefits of interpretability  operating on text embeddings. By prefixing the soft prompts with ``A texture of'', and ``A shape of'', DiffExplainer synthesizes images that maximize the output for the specified class, revealing the textures and shapes most influential in the model's decision-making process.}
  \label{fig:finerunderstanding}
\end{figure}

Operating on text embeddings also provides the flexibility for a nuanced understanding of the model's behavior at a granular level. Indeed, by employing prompts such as ``A texture of'' and generating images that maximize the output for the target class, we gain insight into the textures that most strongly influence the model's decision for that class. Likewise, utilizing shape prompts allows us to identity shapes that significantly impact the target classifier for a given class. Examples demonstrating these capabilities are shown in Fig.~\ref{fig:finerunderstanding}. This approach enables us to delve deeper into the model's decision-making process, shedding light on the features it prioritizes when identifying specific classes.

\subsection{Automated Spurious Feature Discovery}
\label{sec:spurious}

\begin{figure}[t]
  \centering
  \includegraphics[width=1\linewidth]{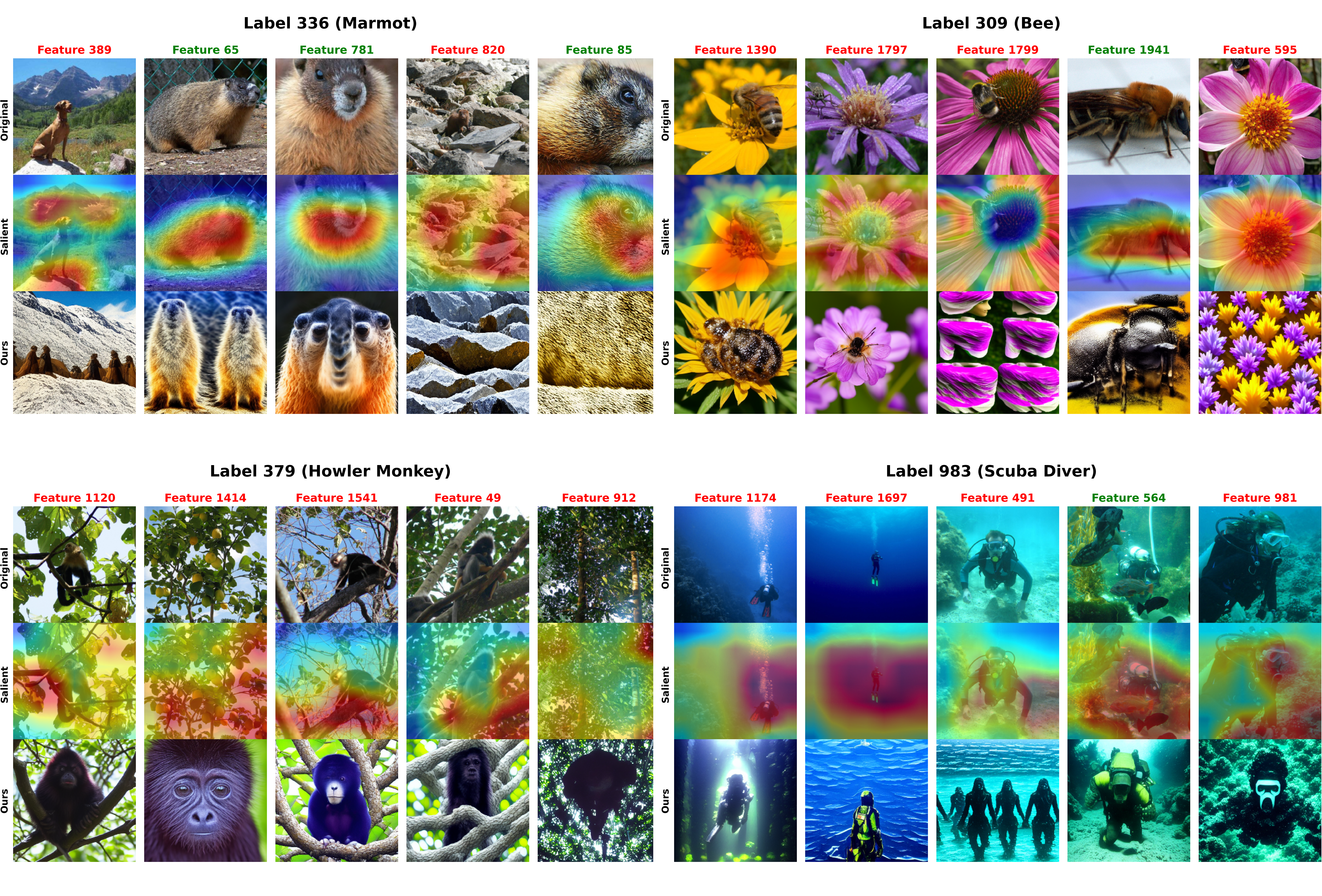}
  \caption{Examples of agreement (top) and disagreement (bottom) between Salient ImageNet annotations and DiffExplainer-generated images. Each block displays the original sample in the first row followed by heatmaps from Salient ImageNet in second row, and images generated by DiffExplainer in the third row. Columns represent neural features: core features (in green) and spurious features (in red) according to Salient ImageNet. 
  }
  \label{fig:salient_qualitative}
\end{figure}

\begin{figure}[htb!]
  \centering
    \includegraphics[width=0.45\textwidth]{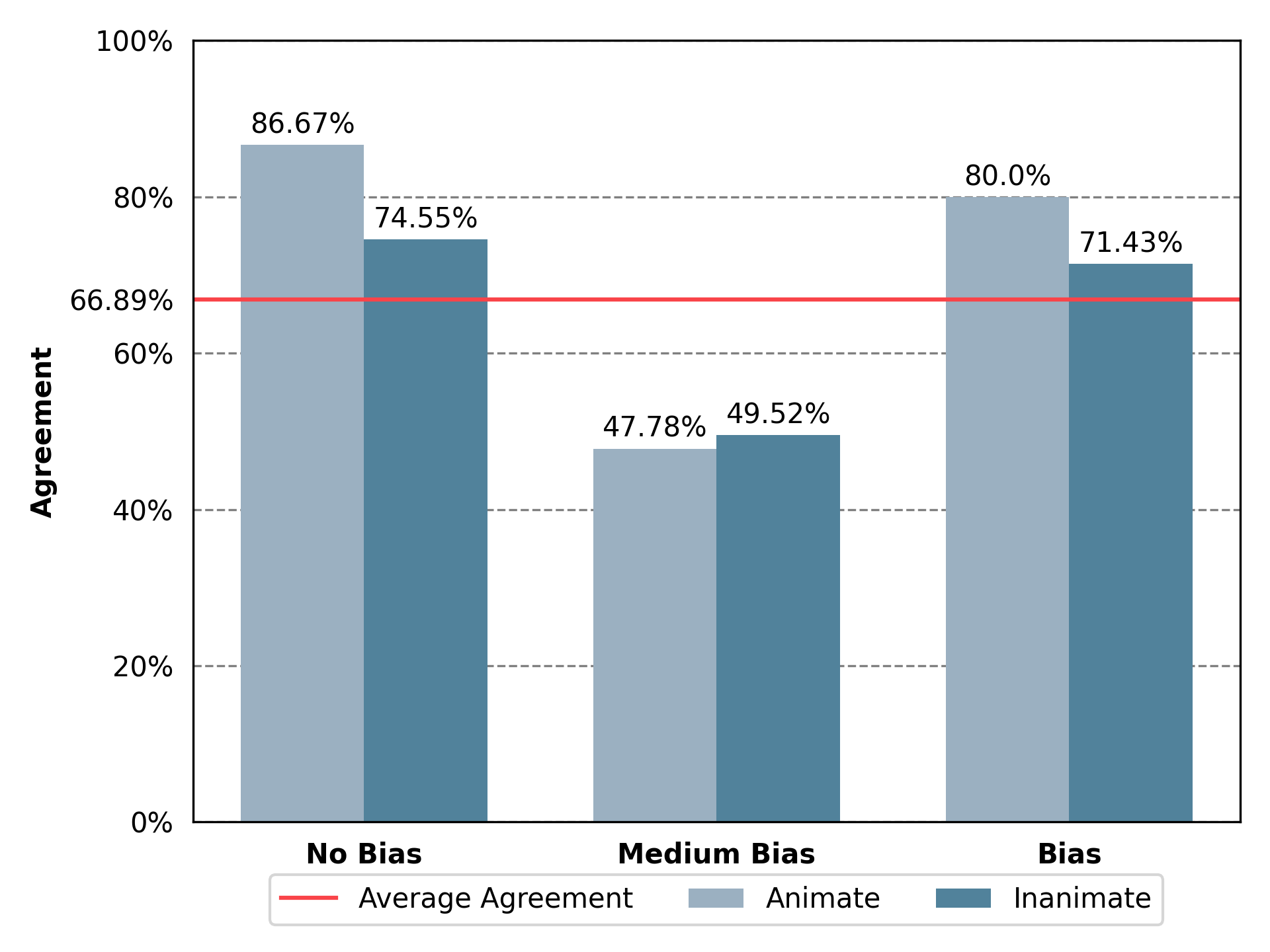}
  \caption{Agreement between DiffExplainer and Salient ImageNet.}
  \label{fig:agreement}
  \end{figure}

We validate our approach in automating the detection of spurious features by comparing our results  with those manually identified in Salient ImageNet.

The first row of Fig.~\ref{fig:salient_qualitative} reports several examples illustrating a significant degree of alignment between our generated images and the classification between core features (in green) and spurious ones (in red). For instance, within the ``Marmot'' class, Salient ImageNet identifies features 389 and 820 as spurious, while features 65, 85, and 781 are recognized as core features. Notably, the GradCAM analyses of features 389 and 820 reveal a focus on rocky areas, which is consistent with the rocky images generated by DiffExplainer. Similarly, images synthesized by DiffExplainer for features 65, 85, and 781 highlight core features for the marmot, consistently matching the activated areas in the Salient ImageNet dataset.

We thus conduct a quantitative evaluation using the methodology outlined in Section~\ref{sec:spurious_discovery} across the 90-class subset of Salient ImageNet, measuring the level of agreement in terms of core/spurious feature identification. The overall agreement between our approach and Salient ImageNet annotations is reported in Fig.~\ref{fig:agreement}.
In details, we categorize the 90 classes into two groups, animate and inanimate objects, and assess the presence of spurious features in Salient ImageNet annotations for each group. Classes with no spurious features are labeled as unbiased, those with all five spurious features are deemed biased, while the rest are considered medium-biased.
Our analysis reveals a high degree of agreement for both biased and unbiased classes, surpassing 80\% for animate objects. Inanimate objects also show a high level of agreement. However, the agreement drops significantly for classes categorized as medium-biased, reaching approximately 50\%.
While this may appear relatively low, it can be attributed to the quality of the Salient ImageNet annotations: these rely on local explanations (e.g., via feature attacks) which may lead to erroneous evaluations from the user study participants.

\subsection{User Study for Feature Preference}
\label{sec:user_study}

\begin{figure}[htb!]
  \centering
    \includegraphics[width=0.55\textwidth]{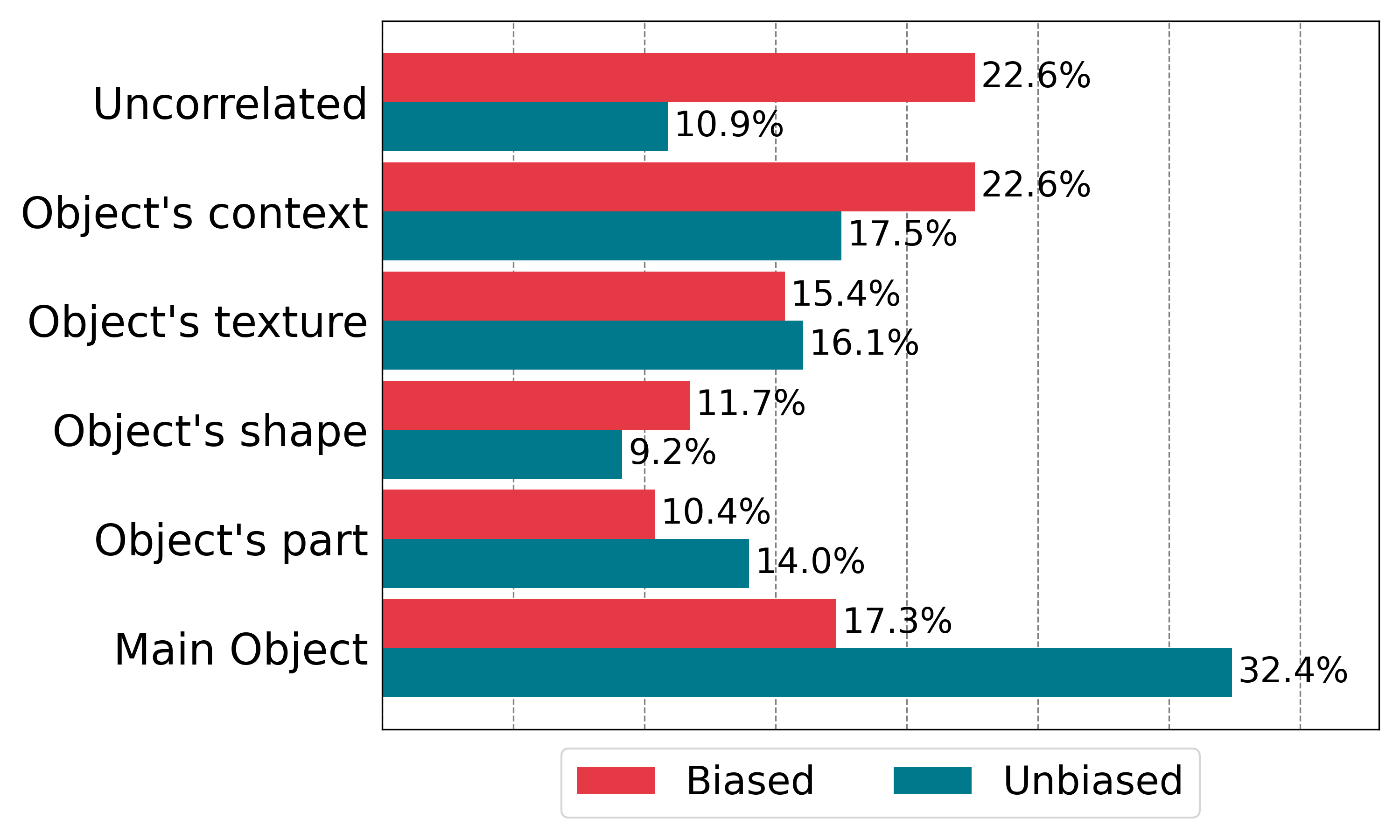}
  \caption{Distribution of features evaluated  by participants in our user study, for classes identified as biased or unbiased by Salient ImageNet.} 
 \label{fig:user_features}
 \end{figure}

While our generated results demonstrate promising outcomes, discrepancies arose between the set of spurious features identified by our method and those presented in Salient ImageNet (see second row of Fig.~\ref{fig:salient_qualitative} and supplementary materials). 
We posit that the cause of these discrepancies lies in the different approaches to defining and explaining bias: namely, a \textit{local} strategy employed by Salient ImageNet as opposed to a \textit{global} approach in our case.

In Salient ImageNet, heatmaps generated via GradCAM are utilized to highlight regions of interest (i.e., areas that mostly activate the classifier output) in images. However, these heatmaps often show multiple visual attributes, complicating the identification of discernible visual patterns. To address this challenge, Salient ImageNet employs feature attack~\cite{Singla_2021_CVPR}, optimizing input images to enhance the activations of desired neural features. Nonetheless, both heatmap and feature attack images constrain human evaluators' ability to identify the features receiving the most attention from the model. In contrast, global explanations provided by DiffExplainer spotlight specific visual patterns in high resolution, enhancing the comprehension of the model's decisions.

To substantiate this claim, we conducted a user study (details in the supplementary materials), involving 100 PhD and MS students in computer science. Participants were asked to evaluate the efficacy of visual features generated by our method versus those generated via feature attack. The targeted classifier is a Robust ResNet-50. 
Specifically, we used soft masks to highlight attributes in real images and asked participants to rate, on a scale from 1 to 5 (1: no resemblance; 5: strong resemblance), the significance of explanation image in relation to the highlighted parts. Images generated by DiffExplainer obtained an average resemblance score of 3.08, while feature attack achieved an average score of 2.40. A t-test established statistical significance in user preference for our approach over feature attack (t-statistic = 10.32, p-value   $\ll$ 0.01).

Furthermore, in Salient ImageNet, context or texture information is often dismissed as spurious features. However, we argue that such features may hold significance when considered alongside core object features. To delve deeper, we asked the same participants to examine images activating both the output neuron and each of the top-5 neural features for a given class, and assess whether the class image depicted the core object and whether the images maximizing the top-5 neural features contained information about the object, its parts, shape, texture, and context.

Fig.~\ref{fig:user_features} illustrates that in the majority of classes identified as biased by Salient ImageNet, images corresponding to the top-5 neural features primarily contained context information, whereas in unbiased classes, these images encompassed information ranging from the entire object (the majority) to texture and context. These results are consistent with our expectations on correct behavior of our method, as well as with the above observation on how Salient ImageNet treats context features. 
\begin{figure}[t]
  \centering
  \includegraphics[width=0.98\linewidth]{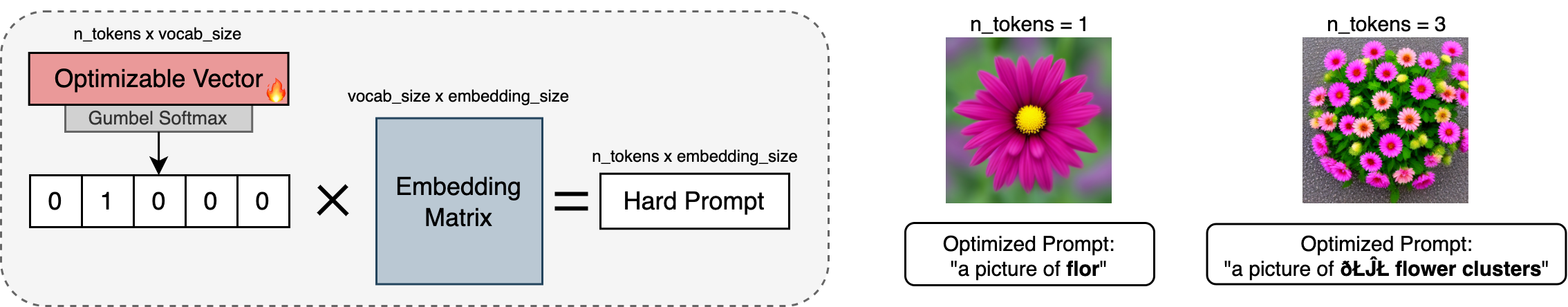}
  \caption{Hard prompt optimization for text explanation in DiffExplainer.}
  \label{fig:hard_prompts}
\end{figure}
\subsection{Towards Text Explanations}
\label{sec:text_expl}
Finally, we report a use case showcasing how operating on text embeddings, as we do in DiffExplainer, opens up the possibility of providing textual explanations by optimizing ``hard prompts'', i.e., ensuring the optimized text embeddings correspond to actual vocabulary prompts. However, replacing soft prompts with  hard prompts is not trivial, as direct optimization in the vocabulary space is not feasible due to its discrete nature. To address this issue, we propose optimizing a continuous vector of the same length as the number of words in the text encoder's embedding layer. While we focus on optimizing just one token for simplicity, the process can be extended to multiple tokens. Subsequently, we apply a Gumbel Softmax~\cite{MaddisonMT17} to approximate it to a one-hot vector. This binary mask is then multiplied with the text encoder embedding matrix, effectively selecting the embedding corresponding to one of the words in the vocabulary, thus making this optimization process reversible to text.\\
A schematic representation of the proposed solution is illustrated in Fig.~\ref{fig:hard_prompts}, with one output example for the ``bee'' class on Robust ResNet-50, providing ``a picture of flower cluster'' as explanation, thus suggesting a bias.

\begin{figure}[t]
  \centering
  \includegraphics[width=1\linewidth]{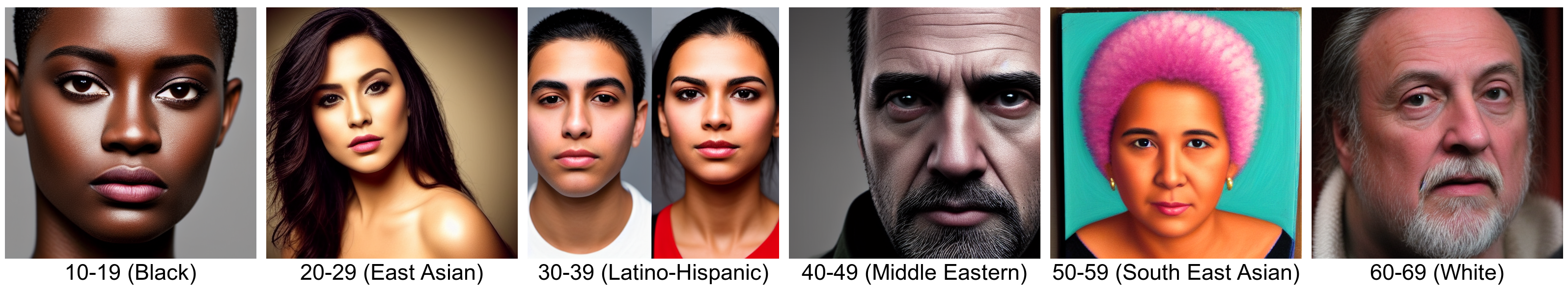}
  \caption{Ethnicity bias identification by DiffExplainer.}
  \label{fig:skin_age_bias}
\end{figure}

\subsection{Ethnicity bias discovery}
\label{sec:gender}
We finally present further qualitative findings on an another benchmark, i.e., FairFaces~\cite{Karkkainen_2021_WACV} dataset. We train a ResNet-50 model for age classification on a modified subset of the dataset with deliberate inclusion of ethnicity biases. Specifically, each age range is associated with a particular ethnicity: 10-19 years with Black, 20-29 with East Asian, 30-39 with Latino-Hispanic, 40-49 with Middle Eastern, 50-59 with South East Asian, and 60-69 with White. We then apply DiffExplainer to the trained model: visual explanations, as depicted in Fig.~\ref{fig:skin_age_bias}, distinctly reveal the injected bias while visualizing fine details, related to the age range, showing its potential in dissecting model decisions.

%% file: sections/conclusions.tex
\section{Conclusions}
\label{sec:conclusions}

Our work has demonstrated the effectiveness and versatility of DiffExplainer in generating high-quality images offering promising opportunities for advancing model explainability. A limitation of our work is the convergence instability (depending on the starting seed) during the image optimization process, that sometimes makes the model stick to local minima, generating unclear images. Additionally, ethical considerations arise due to the potential generation of NSFW images from the latent space of the diffusion model. Future research will focus on improving the optimization strategy addressing both convergence instability and safeguards against unintended content.